\documentclass{article}
\PassOptionsToPackage{numbers, compress}{natbib}
\usepackage[preprint]{main}

\usepackage{times}
\usepackage{microtype}
\usepackage{epsfig}
\usepackage[table,xcdraw]{xcolor}
\usepackage{caption}
\usepackage{float}
\usepackage{placeins}
\usepackage{color, colortbl}
\usepackage{stfloats}
\usepackage{enumitem}
\usepackage{tabularx}
\usepackage{xstring}
\usepackage{multirow}
\usepackage{xspace}
\usepackage{url}
\usepackage{subcaption}
\usepackage{xcolor}
\usepackage[hang,flushmargin]{footmisc}
\usepackage[table]{xcolor}
\usepackage[pagebackref, breaklinks=true, colorlinks, citecolor=citecolor, linkcolor=linkcolor, bookmarks=false]{hyperref}

\usepackage[utf8]{inputenc} 
\usepackage[T1]{fontenc}    
\usepackage{url}            
\usepackage{booktabs}       
\usepackage{amsfonts}       
\usepackage{amsmath}
\usepackage{nicefrac}       
\usepackage{microtype}      
\usepackage[table]{xcolor}  
\usepackage{tabularx}
\usepackage{graphicx}
\usepackage{multirow}
\usepackage{enumitem}
\usepackage{pifont}
\usepackage{adjustbox}

\definecolor{citecolor}{HTML}{0071BC}
\definecolor{linkcolor}{HTML}{ED1C24}
\usepackage[capitalize]{cleveref}
\crefname{section}{Sec.}{Secs.}
\crefname{table}{Table}{Tables}
\crefname{figure}{Fig.}{Figs.}

% ...

\newcommand{\R}[1]{{%
    \textbf{%
        \ifstrequal{#1}{1}{\textcolor{red}{R#1}}{%
        \ifstrequal{#1}{2}{\textcolor{blue}{R#1}}{%
        \ifstrequal{#1}{3}{\textcolor{magenta}{R#1}}{%
        \ifstrequal{#1}{4}{\textcolor{teal}{R#1}}{%
                           \textcolor{cyan}{R#1}%
        }}}}%
    }%
}}

\usepackage{epigraph}
\definecolor{mypurple}{RGB}{200,192,248}
\definecolor{mypurpledeep}{RGB}{142,126,240}
\definecolor{mygreen}{RGB}{117,170,156}
\definecolor{myyellow}{RGB}{255,192,0}
\definecolor{myblue}{RGB}{57,143,255}
\definecolor{mygrey}{RGB}{231,230,230}
\definecolor{codey}{RGB}{220,220,170}
\definecolor{coder}{RGB}{206,145,120}
\definecolor{codeb}{RGB}{156,220,254}
\definecolor{codenum}{RGB}{204,204,204}

\newcommand{\cmark}{\textcolor{mygreen}{\ding{51}}} 

\newcommand{\xmark}{\textcolor{myyellow}{\ding{55}}} 

\usepackage{multirow}
\usepackage[table,xcdraw]{xcolor}
\usepackage[normalem]{ulem}
\useunder{\uline}{\ul}{}

\title{Large Multimodal Agents: A Survey}

\author{
Junlin Xie$^{\clubsuit\heartsuit}$\(^{*}\)\quad
Zhihong Chen$^{\clubsuit\heartsuit}$\thanks{Equal contribution}\quad
Ruifei Zhang$^{\clubsuit\heartsuit}$\quad
Xiang Wan$^\clubsuit$\quad
Guanbin Li$^\spadesuit$\thanks{Corresponding author}\\\\
$^\heartsuit$The Chinese University of Hong Kong, Shenzhen\\$^\clubsuit$Shenzhen Research Institute of Big Data,
$^\spadesuit$Sun Yat-sen University\\
\texttt{\{junlinxie,zhihongchen,ruifeizhang\}@link.cuhk.edu.cn}\\
\texttt{wanxiang@sribd.cn}, \texttt{liguanbin@mail.sysu.edu.cn}}

\begin{document}
\maketitle
\begin{abstract}
Large language models (LLMs) have achieved superior performance in powering text-based AI agents, endowing them with decision-making and reasoning abilities akin to humans.
Concurrently, there is an emerging research trend focused on extending these LLM-powered AI agents into the \underline{\textit{multimodal}} domain. This extension enables AI agents to interpret and respond to diverse multimodal user queries, thereby handling more intricate and nuanced tasks.
In this paper, we conduct a systematic review of LLM-driven multimodal agents, which we refer to as \textit{large multimodal agents} (\texttt{LMAs} for short).
First, we introduce the essential components involved in developing \texttt{LMAs} and categorize the current body of research into four distinct types. Subsequently, we review the collaborative frameworks integrating multiple \texttt{LMAs}, enhancing collective efficacy.
One of the critical challenges in this field is the diverse evaluation methods used across existing studies, hindering effective comparison among different \texttt{LMAs}. Therefore, we compile these evaluation methodologies and establish a comprehensive framework to bridge the gaps. This framework aims to standardize evaluations, facilitating more meaningful comparisons.
Concluding our review, we highlight the extensive applications of \texttt{LMAs} and propose possible future research directions. Our discussion aims to provide valuable insights and guidelines for future research in this rapidly evolving field. An up-to-date resource list is available at \url{https://github.com/jun0wanan/awesome-large-multimodal-agents}.
\end{abstract}
\section{Introduction}
\begin{figure*}[t]
\centering
\includegraphics[width=0.98\linewidth]{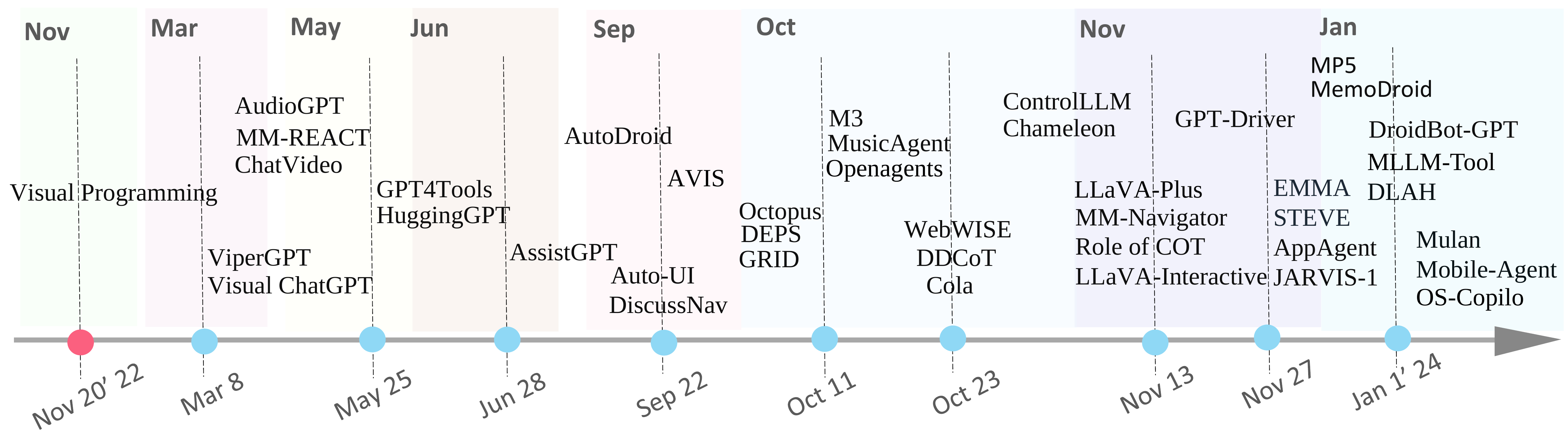}
\caption{Representative research papers from top AI conferences on LLM-powered multimodal agents, published between November 2022 and February 2024, are categorized by model names, with earlier publication dates corresponding to names listed earlier.}
\label{fig:framework1}
\end{figure*}
An \textit{agent} is a system capable of perceiving its \underline{environment} and making \underline{decisions} based on these perceptions to achieve specific \underline{goals} \citep{wooldridge1995intelligent}.
While proficient in narrow domains, early agents\cite{osoba2020policy,wang2020completely} often lack adaptability and generalization, highlighting a significant disparity with human intelligence. Recent advancements in large language models (LLMs) have begun to bridge this gap, where LLMs enhance their capabilities in command interpretation, knowledge assimilation \citep{pan2023large,zhang2023large}, and mimicry of human reasoning and learning \citep{li2022systematic,yang2023supervised}.
These agents use LLMs as their primary decision-making tool and are further enhanced with critical human-like features, such as memory. This enhancement allows them to handle a variety of natural language processing tasks and interact with the environment using language \citep{schick2023toolformer,qin2023toolllm}.

However, real-world scenarios often involve information that spans beyond text, encompassing multiple modalities, with a significant emphasis on the visual aspect. Consequently, the next evolutionary step for LLM-powered intelligent agents is to acquire the capability to process and generate \underline{\textit{multimodal}} information, particularly visual data. This ability is essential for these agents to evolve into more robust AI entities, mirroring human-level intelligence. Agents equipped with this capability are referred to as \textit{large multimodal agents} (\texttt{LMAs}) in our paper.\footnote{The name is inherited from Large Multimodal Models (LMMs).}
Typically, they face more sophisticated challenges than language-only agents. Take web searching for example, an \texttt{LMA} first requires the input of user requirements to look up relevant information through a search bar. Subsequently, it navigates to web pages through mouse clicks and scrolls to browse real-time web page content. Lastly, the \texttt{LMA} needs to process multimodal data (e.g., text, videos, and images) and perform multi-step reasoning, including extracting key information from web articles, video reports, and social media updates, and integrating this information to respond to the user's query.
We note that existing studies in \texttt{LMAs} were conducted in isolation therefore it is necessary to further advance the field by summarizing and comparing existing frameworks.
There exist several surveys related to LLM-powered agents \citep{xi2023rise,sumers2023cognitive,wang2023survey} while few of them focused on the multimodal aspects.

In this paper, we aim to fill the gap by summarizing the main developments of \texttt{LMAs}.
First, we give an introduction about the core components (\S\ref{sec:components}) and propose a new taxonomy for existing studies (\S\ref{sec:taxonomy}) with further discussion on existing collaborative frameworks (\S\ref{sec:multi-agent}). Regarding the evaluation, we outline the existing methodologies for assessing the performance of \texttt{LMAs}, followed by a comprehensive summary (\S\ref{sec:eval}). Then, the application section provides an exhaustive overview of the broad real-world applications of multimodal agents and their related tasks (\S\ref{sec:app}). 
We conclude this work by discussing and suggesting possible future directions for \texttt{LMAs} to provide useful research guidance.
\section{The Core Components of LMAs}
\label{sec:components}
\begin{table*}[t]
    \Large
        \centering
        \begin{adjustbox}{max width=\textwidth}
      \begin{tabular}{llcccccccccccc}
      \toprule
      \multicolumn{1}{c}{\multirow{2}{*}{Type}} & \multicolumn{1}{c}{\multirow{2}{*}{Model}} & \multicolumn{4}{c}{Task Focus} & \multicolumn{4}{c}{Planner} & \multicolumn{2}{c}{Action} & \multirow{2}{*}{Multi-}  & \multirow{2}{*}{Long}\\ \cmidrule(lr){3-6} \cmidrule(lr){7-10}\cmidrule(lr){11-12} 
         & & Text     & Image     & Video &Audio  &Model  & Format        & Inspect    &Planning Method          &Action Type   &Action Learning          &Agent      &Memory       \\ \midrule
          
          \multicolumn{1}{c}{\multirow{21}{*}{Type I}}  
          & VisProg \citep{gupta2023visual}                                   & \cmark  & \cmark   & \xmark & \xmark    &GPT3.5   & Program       & \xmark       &Fix          &T (VFMs \& Python) &Prompt                      &  \xmark      &  \xmark              \\
          & ControlLLM \citep{gupta2023visual}                                   & \cmark  & \cmark   & \xmark & \xmark    &GPT4   & Language       & \cmark       &Fix          &T (VFMs \& Python) &Prompt                      &  \xmark      &  \xmark              \\
          & Visual ChatGPT \citep{wu2023visual}                            & \xmark  & \cmark  & \xmark  & \xmark   &GPT3.5 & Language      & \xmark      &Fix            &T (VFMs)  &Prompt                      &  \xmark        &  \xmark            \\
          & ViperGPT \citep{suris2023vipergpt}                                  & \cmark  & \cmark    & \cmark  & \xmark &GPT3.5 & Program       & \xmark      &Fix            &T (VFMs \& API \& Python)  &Prompt                    &  \xmark        &  \xmark            \\
          & MM-ReAct \citep{yang2023mm}                                  &  \xmark  & \cmark    & \cmark & \xmark &ChatGPT  \&  GPT3.5 & Language      & \cmark     &Fix    &T (Web \& API) &Prompt                   &  \xmark       &  \xmark             \\
          &Chameleon \citep{lu2023chameleon}                                 &  \cmark  & \cmark    & \xmark & \xmark &GPT3.5 & Language      & \xmark      &Fix               &T (VFMs \& API \& Python \& Web) &Prompt                 &  \xmark        &  \xmark            \\
          &HuggingGPT \citep{shen2023hugginggpt}                                &  \xmark & \cmark    & \cmark & \xmark  &GPT3.5 & Language      & \xmark     &Fix                  &T (VFMs) &Prompt                &  \xmark         &  \xmark           \\ 
          &CLOVA \citep{gao2023clova}                       &  \cmark  & \cmark    & \xmark  & \xmark &GPT4 & Language    & \cmark   &Dynamic       &T (VFMs \& API)  &Prompt                        &  \xmark       &  \xmark   \\     
          &CRAFT \citep{yuan2023CRAFT}                       &  \cmark  & \cmark    & \xmark  & \xmark &GPT4 & Program     & \xmark   &Fix       &T (Custom tools)  &Prompt                        &  \xmark       &  \xmark   \\     
          &Cola \citep{chen2023large}                        &  \cmark  & \cmark    & \cmark  & \xmark &ChatGPT & Language     & \xmark   &Fix       &T (VFMs)  &Prompt                        &  \xmark       &  \xmark   \\
          &M3 \citep{liu2023towards}                                        &  \xmark  & \cmark    & \cmark & \xmark &GPT3.5& Language    & \cmark   &Dynamic             &T (VFMs \& API) &Prompt                   &  \xmark             &  \xmark      \\ 
          &DEPS \citep{wang2023describe}       & \cmark  & \cmark    & \xmark & \xmark &GPT-4 &Language &\cmark &Dynamic &E &Prompt &\xmark &\xmark \\
          &GRID  \citep{vemprala2023grid}     & \cmark  & \cmark    & \xmark & \xmark &GPT-4 &Language &\cmark &Dynamic &T (VFMs \& API) &Prompt &\xmark &\xmark \\
          &DroidBot-GPT \citep{wen2023droidbot}       & \cmark  & \cmark    & \xmark & \xmark &ChatGPT &Language &\xmark &Dynamic &V &Prompt &\xmark &\xmark \\
          &ASSISTGUI \citep{gao2023assistgui}      & \cmark  & \cmark    & \xmark & \xmark &GPT-4 &Language &\cmark &Dynamic &V(GUI parser) &Prompt &\xmark &\xmark \\
          &GPT-Driver \citep{mao2023gpt}       & \cmark  & \cmark    & \xmark & \xmark &GPT-3.5 &Language &\xmark &Dynamic &E &Prompt &\xmark &\xmark \\
          &LLaVA-Interactive \citep{chen2023llava}  & \cmark  & \cmark    & \xmark & \xmark &GPT-4 &Language &\xmark &Dynamic &T (VFMs) &Prompt &\xmark &\xmark \\
          &MusicAgent  \citep{yu2023musicagent}       & \cmark  & \xmark    & \xmark & \cmark &ChatGPT &Language &\xmark &Dynamic &T (Music-Models) &Prompt &\xmark &\xmark \\
          &AudioGPT \citep{huang2023audiogpt}       & \cmark  & \xmark    & \xmark & \cmark &GPT-4(V) &Language &\xmark &Fix &T (API) &Prompt &\xmark &\xmark \\
          &AssistGPT \citep{gao2023assistgpt}                                 &  \xmark  & \cmark    & \cmark & \xmark         &GPT3.5 & Lang.  \&  Prog.  & \cmark   &Dynamic             &T (VFMs \& API) &Prompt                     &  \xmark     &  \xmark       
           \\ 
           & Mulan\citep{li2024mulan}     & \cmark  & \cmark   & \xmark & \xmark    &GPT3.5   & Language       & \cmark       &Dynamic          &T (VFMs \& Python) &Prompt                      &  \xmark      &  \xmark              \\
            & Mobile-Agent\citep{ wang2024mobile}     & \cmark  & \cmark   & \xmark & \xmark    &GPT4   & Language       & \cmark       &Dynamic          &V \& T(VFMs) &Prompt                      &  \xmark      &  \xmark              \\
          
\midrule
          \multicolumn{1}{c}{\multirow{8}{*}{Type II}}
           &GPT-Driver \citep{mao2023gpt}                         &  \xmark  & \cmark    & \cmark  & \xmark &GPT4 & Language    & \xmark   &Fix       &E &Learning                        &  \xmark       &  \xmark   \\    
          &LLAVA-PLUS \citep{liu2023llava}                         &  \xmark  & \cmark    & \cmark  & \xmark &Llava& Language    & \cmark   &Dynamic       &T (VFMs) &Learning                        &  \xmark       &  \xmark   \\     
          &GPT4tools \citep{yang2023gpt4tools}                        &  \xmark  & \cmark    & \cmark  & \xmark &Llama & Language    & \cmark   &Dynamic       &T (VFMs) &Learning                        &  \xmark       &  \xmark   \\     
          &Tool-LMM \citep{Wang2024MLLM-Tool}                       &  \cmark  & \cmark    & \cmark  & \cmark &Vicuna & Language    & \xmark   &Dynamic       &T (VFMs \& API) &Learning                        &  \xmark       &  \xmark   \\     
          &STEVE  \citep{zhao2023see}     & \cmark  & \cmark    & \xmark & \xmark &STEVE-13B &Program  &\xmark &Fix &E &Learning &\xmark &\xmark \\
          &EMMA \citep{yang2023embodied}       & \cmark  & \cmark    & \xmark & \xmark &LMDecoder &Language &\xmark &Fix &E &Learning &\xmark &\xmark \\
          &Auto-UI  \citep{zhan2023you}        & \cmark  & \cmark    & \xmark & \xmark &LMDecoder &Language &\xmark &Dynamic &E &Learning &\xmark &\xmark \\
          &WebWISE \citep{tao2023webwise}         & \cmark  & \cmark    & \xmark & \xmark &LMDecoder &Language &\cmark &Dynamic &E &Learning &\xmark &\xmark \\
          
\midrule        
          \multicolumn{1}{c}{\multirow{3}{*}{Type III}}
          &DORAEMONGPT \citep{Yang2024DoraemonGPT}                       &  \cmark  & \xmark    & \cmark  & \xmark &GPT4 & Language    & \xmark   &Dynamic       &T (VFMs)  &Prompt                        &  \xmark       &  \cmark   \\     
          &ChatVideo \citep{wang2023chatvideo}                        &  \cmark  & \cmark    & \cmark  & \xmark &ChatGPT & Language     & \xmark   &Fix       &T (VFMs)  &Prompt                        &  \xmark       &  \cmark   \\ &OS-Copilot \citep{wu2024copilot}                                 &  \cmark  & \cmark    & \xmark & \xmark         &GPT4 & Language  & \cmark   &Dynamic             &V &Prompt                     &  \xmark     &  \cmark \\    
          
\midrule          
          \multicolumn{1}{c}{\multirow{8}{*}{Type IV}}
           &Openagents \citep{xie2023openagents}                                 &  \cmark  & \cmark    & \cmark & \xmark         &GPT3.5 \& GPT4 & Language  & \cmark   &Dynamic             &V \& T &Prompt                     &  \xmark     &  \cmark \\
          &MEIA \citep{liu2024multimodal}                                 &  \cmark  & \cmark    & \xmark & \xmark         &GPT3.5 \& GPT4 & Language  & \xmark   &Fix             &E &Prompt                     &  \xmark     &  \cmark \\
          &JARVIS-1 \citep{wang2023jarvis}  & \cmark  & \cmark    & \xmark & \xmark  &GPT-4  &Language &\cmark &Dynamic  &E &Prompt &\xmark &\cmark \\
          &AppAgent \citep{yang2023appagent}       & \cmark  & \cmark    & \xmark & \xmark &GPT-4(V) &Language &\cmark &Dynamic &E &Prompt &\xmark &\cmark \\
          &MM-Navigator  \citep{yan2023gpt}      & \cmark  & \cmark    & \xmark & \xmark &GPT-4(V) &Language &\xmark &Dynamic &T (API) &Prompt &\xmark &\cmark \\
          &DLAH \citep{fu2023drive}       & \cmark  & \cmark    & \xmark & \xmark &GPT-3.5 &Language &\xmark &Dynamic &V(Simulator-interfaces) &Prompt &\xmark &\cmark \\
          &Copilot  \citep{zhang2023loop}      & \cmark  & \xmark    & \xmark & \cmark &GPT-3.5 &Language &\cmark &Dynamic &T (Music-Models) &Prompt &\xmark &\cmark \\
          &WavJourney \citep{liu2023wavjourney}       & \cmark  & \xmark    & \xmark & \cmark &GPT-4 &Program &\xmark &Dynamic &T (Music-Models) &Prompt &\xmark &\cmark \\
\midrule     
          \multicolumn{1}{c}{\multirow{5}{*}{Multi-agent}}&
           AVIS \citep{hu2023avis}                                      &  \xmark  & \cmark    & \cmark & \xmark &GPT4& Language   & \cmark   &Dynamic             &T (VFMs)  &Prompt                     &  \cmark            &  \xmark       \\ 
     
          & MP5 \citep{qin2023mp5}        & \cmark  & \cmark    & \xmark & \xmark  &GPT-4  &Language &\cmark &Dynamic  &E &Prompt &\cmark &\cmark \\
          & MemoDroid \citep{lee2023explore}       & \cmark  & \cmark    & \xmark & \xmark &GPT-4 &Language &\cmark &Dynamic &V(VFMs \& API) &Prompt &\cmark &\cmark \\

          & DiscussNav \citep{long2023discuss}        & \cmark  & \cmark    & \xmark & \xmark  &GPT-4 \& ChatGPT  &Language &\xmark &Dynamic  &E &Prompt &\cmark &\xmark \\

      \bottomrule
      \end{tabular}
        \end{adjustbox}
        \caption{ 
       This presentation delineates the component details of all \texttt{LMAs}, encompassing their task-specific modalities, the models outlined by planners, the methodologies and formats employed in planning, the variety of actions involved, the extent of multi-agent collaboration, and the incorporation of long-term memory. Within this table, `V' represents the virtual action, `T' indicates the use of a tool, and `E' embodies the physical action.
        }
        \label{tab:ll}
\end{table*}

In this section, we detail four core elements of \texttt{LMAs} including perception, planning, action, and memory.
\paragraph{Perception.}
Perception is a complex cognitive process that enables humans to collect and interpret environmental information. In \texttt{LMAs}, the perception component primarily focuses on processing multimodal information from diverse environments. As illustrated in Table \ref{tab:ll}, \texttt{LMAs} in different tasks involve various modalities. They require extracting key information from these different modalities that is most beneficial for task completion, thereby facilitating more effective planning and execution of the tasks.

Early research \citep{wu2023visual,suris2023vipergpt,yang2023mm,gao2023assistgpt}  on processing multimodal information often rely on simple correlation models or tools to convert images or audio into text descriptions. 
However, this conversion approach tends to generate a large amount of irrelevant and redundant information, particularly for complex modalities (e.g., video). 
Along with the input length constraint, LLMs frequently face challenges in effectively extracting pertinent information for planning. To address this issue, recent studies \citep{Yang2024DoraemonGPT,wang2023chatvideo} have introduced the concept of sub-task tools, which are designed to handle sophisticated data types. 
In an environment resembling the real world (i.e., open-world games), \cite{wang2023jarvis} proposed a novel method for processing non-textual modal information. This approach begins by extracting key visual vocabulary from the environment and then employs the GPT model to further refine this vocabulary into a series of descriptive sentences. As LLMs perceive visual modalities within the environment, they use them to retrieve the most relevant descriptive sentences, which effectively enhances their understanding of the surroundings.

\begin{figure*}[h]
\centering
\includegraphics[width=0.9\textwidth]{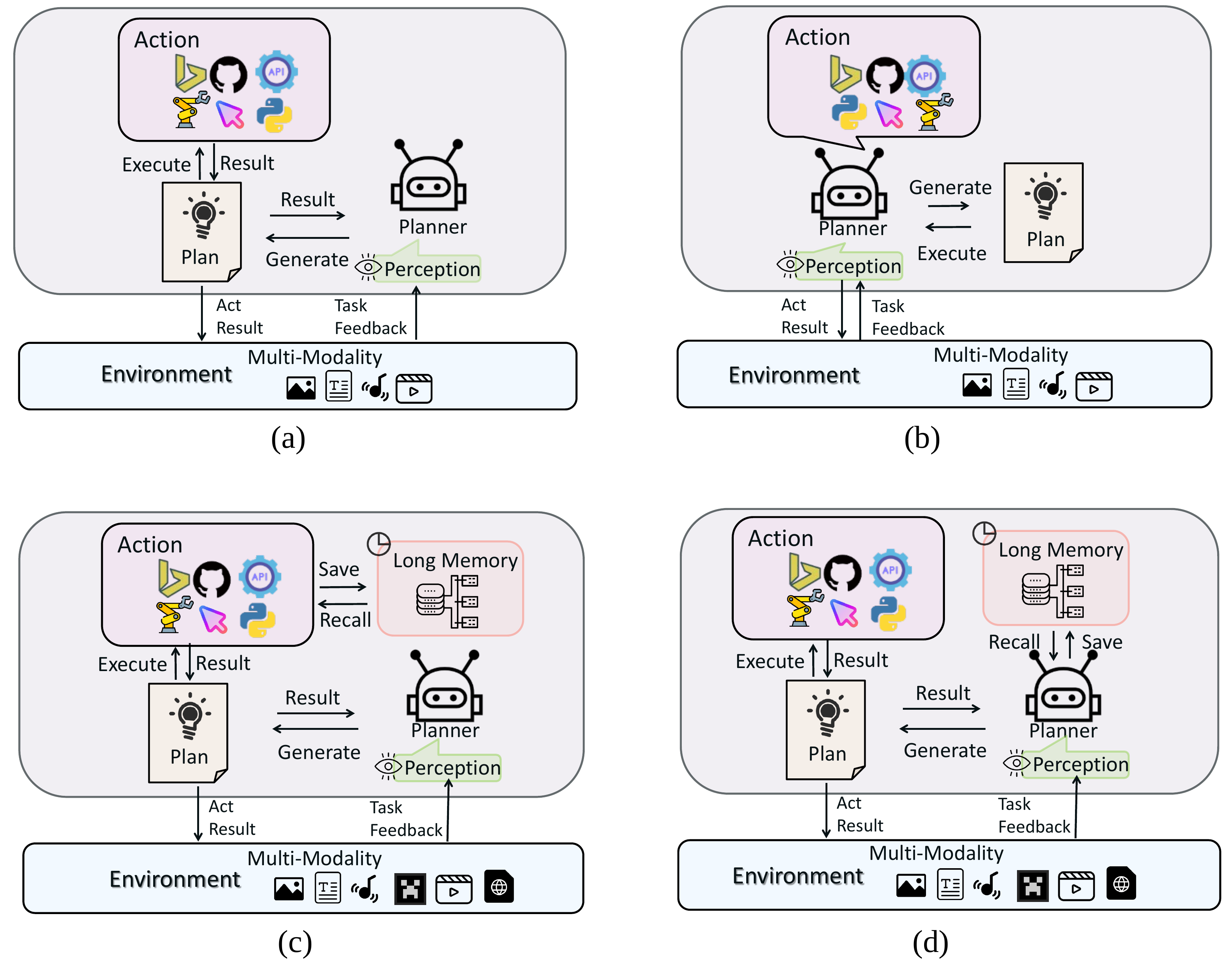}
\caption{Illustrations on four types of \texttt{LMAs}:
(a) Type I: Closed-source LLMs as Planners w/o Long-term Memory. They mainly use prompt techniques to guide closed-source LLMs in decision-making and planning to complete tasks without long memory. (b) Type II:Finetuned LLMs as Planners w/o Long-term Memory. They use action-related data to finetune existing open-source large models, enabling them to achieve decision-making, planning, and tool invocation capabilities comparable to closed-source LLMs. Unlike (a) and (b), (c) and (d) introduce long-term memory functions, further enhancing their generalization and adaptation abilities in environments closer to the real world. However, because their planners use different methods to retrieve memories, they can be further divided into: (c) Type III: Planners with Indirect Long-term Memory; (d) Type IV: Planners with Native Long-term Memory.}
\label{fig:framework-singleagent}
\end{figure*}
\begin{figure*}[h]
\centering
\includegraphics[width=1\textwidth]{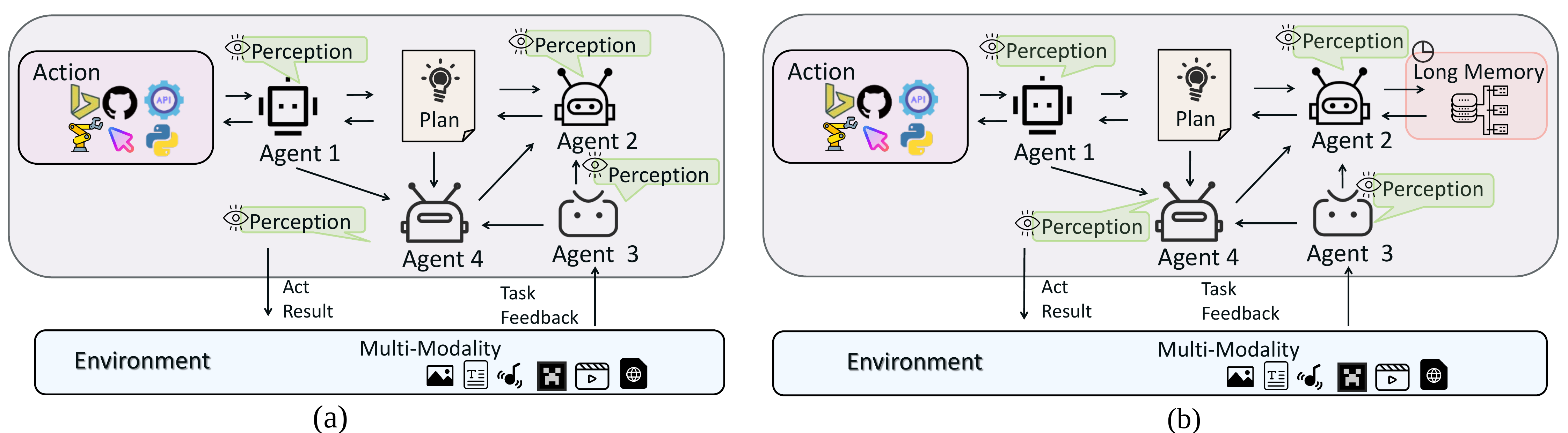}
\caption{Illustrations on two types of multi-agent frameworks: in these two frameworks, facing tasks or instructions from the environment, the completion relies on the cooperation of multiple agents. Each agent is responsible for a specific duty, which may involve processing environmental information or handling decision-making and planning, thus distributing the pressure that would otherwise be borne by a single- agent to complete the task.  The unique aspect of framework (b) is its long-term memory capability.}
\label{fig:framework-multiagent}
\end{figure*}


\paragraph{Planning.} Planners play a central role in \texttt{LMAs}, akin to the function of the human brain. They are responsible for deep reasoning about the current task and formulating corresponding plans. Compared to language-only agents, \texttt{LMAs} operate in a more complicated environment, making it more challenging to devise reasonable plans.
We detail planners from four perspectives (models, format, inspection  \&  reflection, and planning methods):
\begin{itemize}
    \item \textbf{Models}: As shown in Table \ref{tab:ll}, existing studies employ different models as planners. Among them, the most popular ones are GPT-3.5 or GPT-4 \citep{suris2023vipergpt,shen2023hugginggpt,gao2023assistgpt,lu2023chameleon,wu2023visual,wang2023jarvis}. Yet, these models are not publicly available and therefore some studies have begun shifting towards using open-source models, such as LLaMA \citep{yang2023gpt4tools} and LLaVA \citep{liu2023llava}, where the latter can directly process information of multiple modalities, enhancing their ability to make more optimal plans.
    \item \textbf{Format}: It represents how to formulate the plans made by planners. As shown in Table \ref{tab:ll}, there are two formatting ways. The first one is natural language. For example, in \citep{shen2023hugginggpt}, the planning content obtained is ``\textit{The first thing I did was use OpenCV's openpose control model to analyze the pose of the boy in the image....}'', where the plan made is to use ``\textit{OpenCV's openpose control model}''. The second one is in the form of programs, like `` \textit{image\_patch = ImagePatch(image)}'' as described in \citep{suris2023vipergpt}, which invokes the \textit{ImagePatch} function to execute the planning. There are also hybrid forms, such as \citep{gao2023assistgpt}.
    \item \textbf{Inspection  \&  Reflection}: It is challenging for an \texttt{LMAs}  to consistently make meaningful and task-completing plans in a complex multimodal environment. This component aims at enhancing robustness and adaptability. Some research methods \citep{wang2023jarvis,wang2023describe} store successful experiences in long-term memory, including the multimodal states, to guide planning. During the planning process, they first retrieve relevant experiences, aiding planners in thoughtful deliberation to reduce uncertainty. Additionally, \cite{hu2023avis} utilizes plans made by humans in different states while performing the same tasks. When encountering similar states, planners can refer to these ``standard answers" for contemplation, leading to more rational plans. Moreover, \cite{Yang2024DoraemonGPT} employs more complex planning methods, like Monte Carlo, to expand the scope of planning search to find the optimal planning strategy.
    \item \textbf{Planning Methods}: Existing planning strategies can be categorized into two types: dynamic and static planning as shown in Table \ref{tab:ll}. The former \citep{wu2023visual,suris2023vipergpt,yang2023mm,lu2023chameleon,shen2023hugginggpt} refers to decomposing the goal into a series of sub-plans based on the initial input, similar to Chain of Thought (CoT) \citep{zheng2023ddcot}, where plans are not reformulated even if errors occur during the process; The latter \citep{gao2023assistgpt,liu2023towards,wang2023jarvis,Yang2024DoraemonGPT} implies that each plan is formulated based on the current environmental information or feedback. If errors are detected in the plan, it will revert to the original state for re-planning \citep{hu2023avis}.
\end{itemize}

\paragraph{Action.} The action component in multimodal agent systems is responsible for executing the plans and decisions formulated by the planner. It translates these plans into specific actions, such as the use of tools, physical movements, or interactions with interfaces, thereby ensuring that the agent can achieve its goals and interact with the environment accurately and efficiently. Our discussion focuses on two aspects: types and approaches.

Actions in Table \ref{tab:ll} are classified into three categories: tool use (T), embodied actions (E), and virtual actions (V), where tools includes visual foundation models (VFMs), APIs, Python, etc (as listed in Table~\ref{tab:tools}); Embodied actions are performed by physical entities like robots \citep{mao2023gpt,fu2023drive} or virtual characters \citep{wang2023jarvis,wang2023describe,vemprala2023grid,yang2023embodied}; Virtual actions \citep{gao2023assistgui,zhang2023bootstrap,tao2023webwise,wen2023droidbot} include web tasks (e.g., clicking links, scrolling, and keyboard use).
In terms of approaches, as shown in Table \ref{tab:ll}, there are primarily two types. The first type involves using prompts to provide agents with information about executable actions, such as the tools available at the moment and their functions; The second type involves collecting data on actions and leveraging this information to self-instruct the fine-tuning process of open-source large models, such as LLaVA \citep{liu2023llava}. This data is typically generated by advanced models, such as GPT-4. Compared to language-only agents, the complexity of information and data related to actions requires more sophisticated methods to optimize the learning strategy.



\begin{table}[htbp]
  \centering
  \begin{adjustbox}{max width=\textwidth}
  \begin{tabular}{llll}
    \toprule
    Modality & Skill & Tools & Source \\
    \midrule
    \multirow{7}{*}{Image}   & VQA  & BLIP2 \citep{li2023BLIP}   & Github, HuggingFace   \\
       & Grounding/Detection   & G-DINO \citep{liu2023grounding}   & Github, HuggingFace   \\
       & Image Caption   & BLIP \citep{li2022BLIP},BLIP2 \citep{li2023BLIP},InstructBLIP \citep{dai2305instructBLIP}   & Github, HuggingFace, API   \\
       & OCR   & EasyOCR,Umi-OCR   & Github, API   \\
       & Image Editing   & Instruct P2P \citep{xu2023instructp2p}   & Github, HuggingFace, API   \\
       & Image Generation   & Stable Diffusion \citep{rombach2022high}, DALLE·3 \citep{betker2023improving}   & Github, HuggingFace, API   \\
        & Image Segmentation   & SAM \citep{kirillov2023segment}, PaddleSeg \citep{liu2021paddleseg}  & Github, HuggingFace, API   \\\midrule
       \multirow{2}{*}{Text} & Knowledge Retrieval   & Bing Search  & Website, API   \\
        & Programming Related Skill   & PyLint, PyChecker   & Python, API   \\\midrule
       \multirow{2}{*}{Video} & Video Editing   & Editly   & Github, API   \\
         & Object-tracking   & OSTrack \citep{ye2022joint}  & Github, HuggingFace, API   \\\midrule
        \multirow{2}{*}{Audio} & Speech to Text  & Whisper \citep{cao2012whisper} & Github, HuggingFace, API   \\
         & Text to Speech   & StyleTTS 2 \citep{li2024styletts} & Github, API   \\
    \bottomrule
  \end{tabular}
  \end{adjustbox}
  \vspace{1em}
\caption{A summary of different tools, including their corresponding modalities, skills, and available sources.}
  \label{tab:tools}
\end{table}

\paragraph{Memory.} Early studies show that memory mechanisms play a vital role in the operation of general-purpose agents. Similar to humans, memory in agents can be categorized into long and short memory. In a simple environment, short memory suffices for an agent to handle tasks at hand. However, in more complex and realistic settings, long memory becomes essential. 
In Table \ref{tab:ll}, we can see that only a minority of \texttt{LMAs} incorporate long memory. Unlike language-only agents, these multimodal agents require long memory capable of storing information across various modalities. 
In some studies \citep{Yang2024DoraemonGPT,wang2023chatvideo,yang2023appagent,fu2023drive}, all modalities are converted into textual formats for storage. However, in \cite{wang2023jarvis}, a multimodal long memory system is proposed, designed specifically to archive previous successful experiences. Specifically, these memories are stored as key-value pairs, where the key is the multimodal state and the value is the successful plan. Upon encountering a new multimodal state, the most analogous examples are retrieved based on their encoded similarity:
\begin{equation}
    p(t|x) \propto \text{CLIP}_v(k_t)^\top \text{CLIP}_v(k_x),
\end{equation}
where \( k_t \) represents the key's visual information encoded via the CLIP model, compared for similarity with the current visual state \( k_x \), also encoded by CLIP.

\section{The Taxonomy of LMAs}
\label{sec:taxonomy}
In this section, we present a taxonomy of existing studies by classifying them into four types.

\paragraph{Type I: Closed-source LLMs as Planners w/o Long-term Memory.} Early studies \citep{gupta2023visual,suris2023vipergpt,wu2023visual,shen2023hugginggpt,gao2023assistgpt,liu2023towards} employ prompts to utilize closed-source large language models (e.g., GPT-3.5) as the planner for inference and planning as illustrated in Figure \ref{fig:framework-singleagent}(a). 
Depending on the specific environment or task requirements, the execution of these plans may be carried out by downstream toolkits or through direct interaction with the environment using physical devices like mice or robotic arms. \texttt{LMAs} of this type typically operate in simpler settings, undertaking conventional tasks such as image editing, visual grounding, and visual question answering (VQA).

\paragraph{Type II: Finetuned LLMs as Planners w/o Long-term Memory.} \texttt{LMAs} of this type involve collecting multimodal instruction-following data or employing self-instruction to fine-tune open-source large language models (such as LLaMA) \citep{yang2023gpt4tools} or multimodal models (like LLaVA) \citep{liu2023llava,Wang2024MLLM-Tool}, as illustrated in Figure \ref{fig:framework-singleagent}(b).  This enhancement not only allows the models to serve as the central ``brain" for reasoning and planning but also to execute these plans. The environments and tasks faced by Type II \texttt{LMAs} are similar to those in Type I, typically involving traditional visual or multimodal tasks.
Compared to canonical scenarios characterized by relatively simple dynamics, closed environments, and basic tasks, \texttt{LMAs} in open-world games like Minecraft are required to execute precise planning in dynamic contexts, handle tasks of high complexity, and engage in lifelong learning to adapt to new challenges. Therefore, building upon the foundation of Type I and Type II, Type III and Type IV \texttt{LMAs} integrate a memory component, showing great promise in developing towards a generalist agent in the field of artificial intelligence.

\paragraph{Type III: Planners with Indirect Long-term Memory.} For Type III \texttt{LMAs} \citep{Yang2024DoraemonGPT,wang2023chatvideo}, as illustrated in Figure \ref{fig:framework-singleagent}(c), LLMs function as the central planner and are equipped with long memory. These planners access and retrieve long memories by invoking relevant tools, leveraging these memories for enhanced reasoning and planning.
For example, the multimodal agent framework developed in \citep{Yang2024DoraemonGPT} is tailored for dynamic tasks such as video processing. This framework consists of a planner, a toolkit, and a task-relevant memory bank that catalogues spatial and temporal attributes. The planner employs specialized sub-task tools to query the memory bank for spatiotemporal attributes related to the video content, enabling inference on task-relevant temporal and spatial data. Stored within the toolkit, each tool is designed for specific types of spatiotemporal reasoning and acts as an executor within the framework. 

\paragraph{Type IV: Planners with Native Long-term Memory.} Different from Type III, Type IV \texttt{LMAs} \citep{wang2023jarvis,qin2023mp5,fu2023drive,zhang2023bootstrap} feature LLMs directly interacting with long memory, bypassing the need for tools to access long memories, as illustrated  in Figure \ref{fig:framework-singleagent}(d). For example, the multimodal agent proposed in \cite{wang2023jarvis} demonstrates proficiency in completing over 200 distinct tasks within the open-world context of Minecraft. In their multimodal agent design, the interactive planner, merging a multimodal foundation model with an LLM, first translates environmental multimodal inputs into text. 
The planner further employs a self-check mechanism to anticipate and assess each step in execution, proactively spotting potential flaws and, combined with environmental feedback and self-explanation, swiftly corrects and refines plans without extra information. Moreover, this multimodal agent framework includes a novel multimodal memory. Successful task plans and their initial multimodal states are stored, and the planner retrieves similar states from this database for new tasks, using accumulated experiences for faster, more efficient task completion.
\section{Multi-agent Collaboration}
\label{sec:multi-agent}
We further introduce the collaborative framework for \texttt{LMAs} beyond the discussion within isolated agents in this section.

As shown in Figure \ref{fig:framework-multiagent}(a)(b), these frameworks employ multiple \texttt{LMAs} working collaboratively. The key distinction between the two frameworks lies in the presence or absence of a memory component, but their underlying principle is consistent: 
multiple \texttt{LMAs} have different roles and responsibilities, enabling them to coordinate actions to collectively achieve a common goal. This structure alleviates the burden on a single agent, thereby enhancing task performance \citep{hu2023avis,qin2023mp5,lee2023explore,long2023discuss}.

For example, in Table \ref{tab:ll}, in the multimodal agent framework by \cite{qin2023mp5}, a perceiver agent is introduced to sense the multimodal environment, comprised of large multimodal models. An agent, designated as Patroller, is responsible for engaging in multiple interactions with the perceiver agent, conducting real-time checks and feedback on the perceived environmental data to ensure the accuracy of current plans and actions. When execution failures are detected or reevaluation is necessitated, Patroller provides pertinent information to the planner, prompting a reorganization or update of the action sequences under the sub-goals.
The MemoDroid framework \citep{lee2023explore} comprises several key agents that collaboratively work to automate mobile tasks. The Exploration Agent is responsible for offline analysis of the target application interface, generating a list of potential sub-tasks based on UI elements, which are then stored in the application memory. During the online execution phase, the Selection Agent determines specific sub-tasks to execute from the explored set, based on user commands and the current screen state. The Deduction Agent further identifies and completes the underlying action sequences required for the selected sub-tasks by prompting an LLM. Concurrently, the Recall Agent, upon encountering tasks similar to those previously learned, can directly invoke and execute the corresponding sub-tasks and action sequences from memory.
\section{Evaluation}
\label{sec:eval}
The predominant focus of research is on enhancing the capabilities of current \texttt{LMAs}. However, limited efforts are devoted to developing methodologies for the assessment and evaluation of these agents. The majority of research continues to depend on conventional metrics for evaluating performance, clearly illustrating the challenges inherent in assessing \texttt{LMAs}. This also underscores the necessity of developing pragmatic assessment criteria and establishing benchmark datasets in this domain. 
This section summarizes existing evaluations of \texttt{LMAs} and offers perspectives on future developments.

\subsection{Subjective Evaluation}
Subjective Evaluation mainly refers to using humans to assess the capabilities of these \texttt{LMAs}.
Our ultimate goal is to create a \texttt{LMA} that can comprehend the world like humans and autonomously execute a variety of tasks. 
Therefore, it is crucial to adopt subjective evaluations of human users on the capabilities of \texttt{LMAs}. The main evaluation metrics include versatility, user-friendliness, scalability, and value and safety.

\paragraph{Versatility.}
Versatility denotes the capacity of an \texttt{LMA} to adeptly utilize diverse tools, execute both physical and virtual actions, and manage assorted tasks. \cite{lu2023chameleon} propose comparing the scale and types of tools utilized in existing \texttt{LMAs}, as well as assessing the diversity of their capabilities.

\paragraph{User-Friendliness.}
User-friendliness involves user satisfaction with the outcomes of tasks completed by LMAs, including efficiency, accuracy, and the richness of the results. 
This type of assessment is relatively subjective. In \cite{yan2023gpt}, human evaluation of the \texttt{LMA} is essential to precisely assess its effectiveness in interpreting and executing user instructions. 

\paragraph{Scalability.}
Scalability fundamentally evaluates the capability of \texttt{LMAs} to assimilate new competencies and address emerging challenges. Given the dynamic nature of human requirements, it is imperative to rigorously assess the adaptability and lifelong learning potential of LMAs.  For example, the evaluation in \cite{liu2023llava} focuses on the proficiency of agents in using previously unseen tools to complete tasks.

\paragraph{Value and Safety.} 
In addition to the metrics previously mentioned, the ``Value and Safety'' metric plays a critical role in determining the practical significance and safety of agents for human users. While many current evaluations overlook this metric, it is essential to consider the ``Value and Safety'' of \texttt{LMAs}. Compared to language agents, \texttt{LMAs} can handle a wider range of task categories, making it even more important for them to follow ethical and moral principles consistent with human societal values.

\subsection{Objective Evaluation}
Objective evaluation, distinct from subjective assessment, relies on quantitative metrics to comprehensively, systematically, and standardizedly assess the capabilities of \texttt{LMAs}. It is currently the most widely adopted evaluation method in multimodal agent research.

\paragraph{Metrics.}
Metrics play a crucial role in objective assessment.  In current multimodal agent research \citep{suris2023vipergpt,yang2023mm,gao2023assistgpt,Yang2024DoraemonGPT,hu2023avis,wu2023visual,lu2023chameleon}, specific task-related metrics are employed, such as the accuracy of answers generated by the agent in tasks like visual question answering (VQA) \citep{gao2023clova,suris2023vipergpt}. However, the traditional task metrics established prior to the emergence of LLMs are not sufficiently effective in evaluating llm-powered \texttt{LMAs}. As a result, an increasing number of research efforts are directed towards identifying more appropriate metrics for assessment. For instance, in VisualWebArena \citep{Koh2024VisualWebArena}, a specialized assessment metric is designed to evaluate the performance of \texttt{LMAs} in handling visually guided tasks. This includes measuring the accuracy of the agent's visual understanding of webpage content, such as the ability to recognize and utilize interactable elements marked by Set-of-Marks for operations and achieving state transitions based on task objectives, as defined by a manually designed reward function. Besides, it encompasses the accuracy of responses to specific visual scene questions and the alignment of actions executed based on visual information.

\paragraph{Benchmarks.}
Benchmark represents a testing environment that encompasses a suite of evaluation standards, datasets, and tasks. It is utilized to assess and compare the performance of different algorithms or systems. Compared to benchmarks for conventional tasks \citep{lu2023chameleon,hu2023avis,wu2023visual,liu2023llava}, SmartPlay \citep{wu2023smartplay} utilizes a carefully designed set of games to comprehensively measure the various abilities of \texttt{LMAs}, establishing detailed evaluation metrics and challenge levels for each capability. 
Contrasting with the approach of using games to evaluate, GAIA \citep{mialon2023gaia} has developed a test set comprising 466 questions and their answers. These questions require AI systems to possess a range of fundamental abilities, such as reasoning, processing multimodal information, web navigation, and proficient tool use. 
Diverging from the current trend of creating increasingly difficult tasks for humans, it focuses on conceptually simple yet challenging questions for existing advanced AI systems. These questions involve real-world scenarios that necessitate the precise execution of complex operational sequences, with outputs that are easy to verify. 
Similarly, VisualWebArena \citep{Koh2024VisualWebArena} is a benchmark test suite designed to assess and advance the capabilities of \texttt{LMAs} in processing visual and textual understanding tasks on real webpages. There are also other benchmarks\citep{lu2024weblinx,xie2024travelplanner} that have effectively tested the capabilities of agents.

\section{Application} 
\label{sec:app}
\begin{figure}[t]
\centering
\includegraphics[width=.98\textwidth]{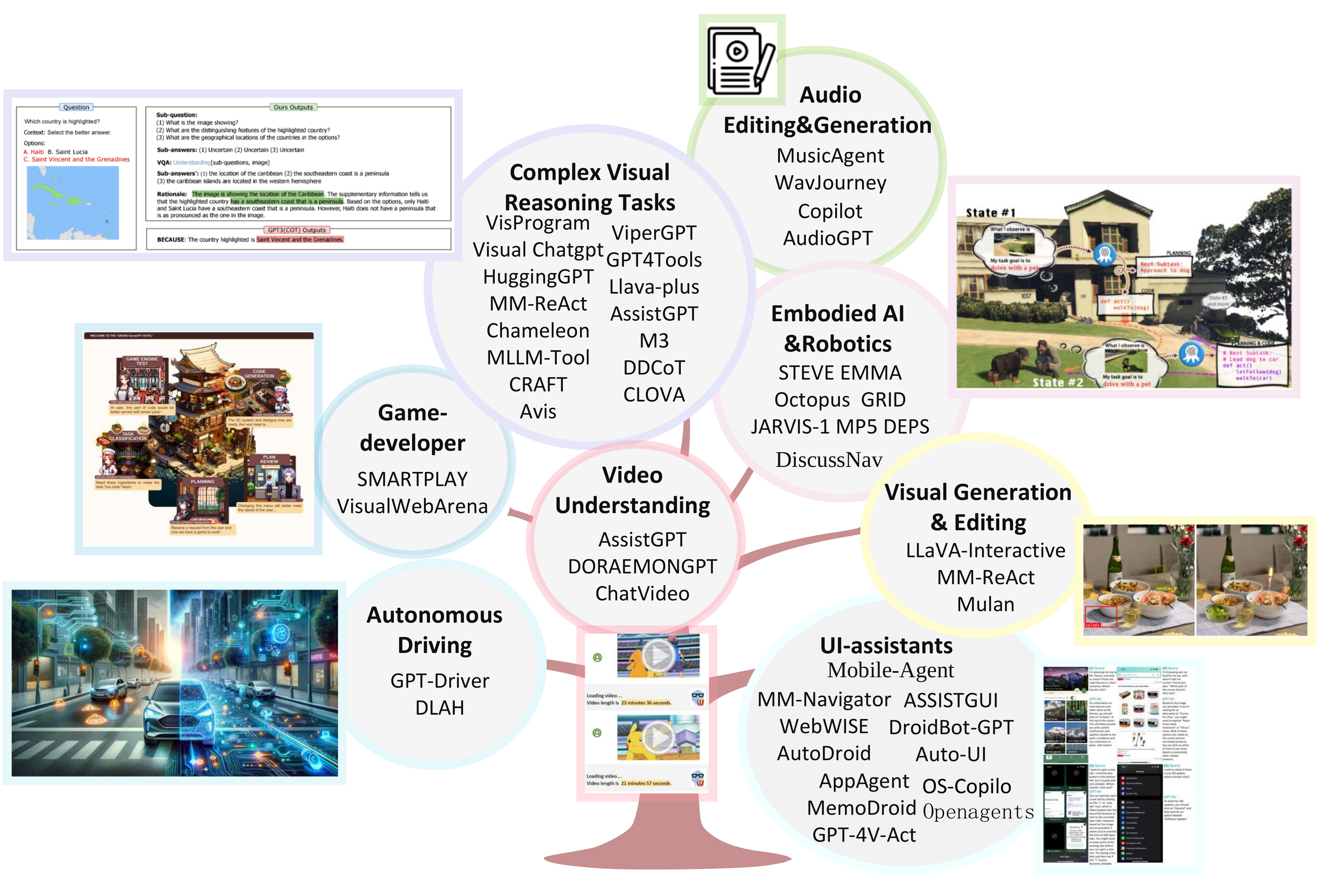}
\caption{A variety of applications of \texttt{LMAs}.}
\label{fig:app}
\vspace{-0.5cm}
\end{figure}

\texttt{LMAs}, proficient in processing diverse data modalities, surpass language-only agents in decision-making and response generation across varied scenarios. Their adaptability makes them exceptionally useful in real-world, multisensory environments, as illustrated in Figure \ref{fig:app}.

\paragraph{GUI Automation.} In this application, the objective of \texttt{LMAs} is to understand and simulate human actions within user interfaces, enabling the execution of repetitive tasks, navigation across multiple applications, and the simplification of complex workflows. This automation holds the potential to save users' time and energy, allowing them to focus on the more critical and creative aspects of their work  \citep{tao2023webwise,ddupont808:2023:GPT-4V-Act,wen2023empowering,yan2023gpt,zhan2023you,yang2023appagent,wen2023droidbot,lee2023explore,gao2023assistgui}.
For example, GPT-4V-Act \citep{ddupont808:2023:GPT-4V-Act}, is an advanced AI that combines GPT-4V's capabilities with web browsing to improve human-computer interactions. Its main goal is to make user interfaces more accessible, simplify workflow automation, and enhance automated UI testing. This AI is especially beneficial for people with disabilities or limited tech skills, helping them navigate complex interfaces more easily.

\paragraph{Robotics and Embodied AI.}  This application \citep{qin2023mp5,wang2023jarvis,yang2023embodied,wang2023describe,vemprala2023grid,yang2023octopus,zhao2023see} focuses on integrating the perceptual, reasoning, and action capabilities of robots with physical interactions in their environments. Employing a multimodal agent, robots are enabled to utilize diverse sensory channels, such as vision, audition, and touch, to acquire comprehensive environmental data. For example, the MP5 system \citep{qin2023mp5}, is a cutting-edge, multimodal entity system used in Minecraft that utilizes active perception to smartly break down and carry out extensive, indefinite tasks with large language models.

\paragraph{Game Developement.} Game AI \citep{wu2023smartplay,Koh2024VisualWebArena} endeavors to design and implement these agents to exhibit intelligence and realism, thereby providing engaging and challenging player experiences. The successful integration of agent technology in games has led to the creation of more sophisticated and interactive virtual environments. 

\paragraph{Autonomous Driving.} Traditional approaches to autonomous vehicles \citep{maurer2016autonomous} face obstacles in effectively perceiving and interpreting complex scenarios. Recent progress in multimodal agent-based technologies, notably driven by LLMs, marks a substantial advancement in overcoming these challenges and bridging the perception gap \citep{mao2023gpt,fu2023drive,zhou2023vision,wen2023road}. \cite{mao2023gpt} present GPT-Driver, a pioneering approach that employs the OpenAI GPT-3.5 model as a reliable motion planner for autonomous vehicles, with a specific focus on generating safe and comfortable driving trajectories. Harnessing the inherent reasoning capabilities of LLMs, their method provides a promising solution to the issue of limited generalization in novel driving scenarios.

\paragraph{Video Understanding.} The video understanding agents \citep{gao2023assistgpt,Yang2024DoraemonGPT} are artificial intelligence systems specifically designed for analyzing and comprehending video content. It utilizes deep learning techniques to extract essential information from videos, identifying objects, actions, and scenes to enhance understanding of the video content. 

\paragraph{Visual Generation \& Editing.} Applications of this kind \citep{chen2023llava,yang2023mm,wang2023chatvideo} are designed for the creation and manipulation of visual content. Using advanced technologies, this tool effortlessly creates and modifies images, offering users a flexible option for creative projects. For instance, LLaVA-Interactive \citep{chen2023llava} is an open-source multimodal interactive system that amalgamates the capabilities of pre-trained AI models to facilitate multi-turn dialogues with visual cues and generate edited images, thereby realizing a cost-effective, flexible, and intuitive AI-assisted visual content creation experience.

\paragraph{Complex Visual Reasoning Tasks.} 
This area is a key focus in multimodal agent research, mainly emphasizing the analysis of multimodal content. This prevalence is attributed to the superior cognitive capabilities of LLMs in comprehending and reasoning through knowledge-based queries, surpassing the capabilities of previous models \citep{karthik2023vision,liu2023towards,zheng2023ddcot}. Within these applications, the primary focus is on QA tasks \citep{shen2023hugginggpt,wu2023visual,yang2023mm,lu2023chameleon}. This entails leveraging visual modalities (images or videos) and textual modalities (questions or questions with accompanying documents) for reasoned responses. 

\paragraph{Audio Editing \& Generation.} The \texttt{LMAs} in this application integrate foundational expert models in the audio domain, making the editing and creation of music efficient\citep{zhang2023loop,yu2023musicagent,huang2023audiogpt,liu2023wavjourney}.
\section{Conclusion and Future Research}
In this survey, we provide a thorough overview of the latest research on multimodal agents driven by LLMs (\texttt{LMAs}). 
We start by introducing the core components of \texttt{LMAs} (i.e., perception, planning, action, and memory) and classify existing studies into four categories. Subsequently, we compile existing methodologies for evaluating \texttt{LMAs} and devise a comprehensive evaluation framework. Finally, we spotlight a range of current and significant application scenarios within the realm of \texttt{LMAs}.
Despite the notable progress, this field still faces many unresolved challenges, and there is considerable room for improvement. We finally highlight several promising directions based on the reviewed progress: 
\begin{itemize}
    \item On frameworks: The future frameworks of \texttt{LMAs} may evolve from two distinct perspectives. From the viewpoint of a single agent, development could progress towards the creation of a more \underline{unified} system. This entails planners directly interacting with multimodal environments \citep{Yang2024DoraemonGPT}, utilizing a comprehensive set of tools \citep{lu2023chameleon}, and manipulating memory directly \citep{wang2023jarvis}; From the perspective of multiple agents, advancing the effective coordination among multiple multimodal agents for the execution of collective tasks emerges as a critical research trajectory. This encompasses essential aspects such as collaborative mechanisms, communication protocols, and strategic task distribution.
    \item On evaluation: Systematic and standard evaluation frameworks are highly desired for this field. An ideal evaluation framework should encompass a spectrum of assessment tasks  \citep{wu2023smartplay,Koh2024VisualWebArena}, varying from straightforward to intricate, each bearing significant relevance and utility for humans. It ought to incorporate lucid and judicious evaluation metrics, meticulously designed to evaluate the diverse capabilities of an \texttt{LMA} in a comprehensive, yet non-repetitive manner. Moreover, the dataset used for evaluation should be meticulously curated to reflect a closer resemblance to real-world scenarios.
    \item On application: The potential applications of \texttt{LMAs} in the real world are substantial, offering solutions to problems that were previously challenging for conventional models, such as web browsing. Furthermore, the intersection of \texttt{LMAs} with the field of human-computer interaction \citep{wen2023droidbot,tao2023webwise} represents one of the significant directions for future applications. Their ability to process and understand information from various modalities enables them to perform more complex and nuanced tasks, thereby enhancing their utility in real-world scenarios and improving the interaction between humans and machines.
\end{itemize}
\clearpage
{\small \bibliographystyle{plain} \bibliography{main}}

\begin{thebibliography}{10}

\bibitem{betker2023improving}
James Betker, Gabriel Goh, Li~Jing, Tim Brooks, Jianfeng Wang, Linjie Li, Long Ouyang, Juntang Zhuang, Joyce Lee, Yufei Guo, et~al.
\newblock Improving image generation with better captions.
\newblock {\em Computer Science. https://cdn. openai. com/papers/dall-e-3. pdf}, 2:3, 2023.

\bibitem{cao2012whisper}
Nan Cao, Yu-Ru Lin, Xiaohua Sun, David Lazer, Shixia Liu, and Huamin Qu.
\newblock Whisper: Tracing the spatiotemporal process of information diffusion in real time.
\newblock {\em IEEE transactions on visualization and computer graphics}, 18(12):2649--2658, 2012.

\bibitem{chen2023large}
Liangyu Chen, Bo~Li, Sheng Shen, Jingkang Yang, Chunyuan Li, Kurt Keutzer, Trevor Darrell, and Ziwei Liu.
\newblock Large language models are visual reasoning coordinators.
\newblock {\em arXiv preprint arXiv:2310.15166}, 2023.

\bibitem{chen2023llava}
Wei-Ge Chen, Irina Spiridonova, Jianwei Yang, Jianfeng Gao, and Chunyuan Li.
\newblock Llava-interactive: An all-in-one demo for image chat, segmentation, generation and editing.
\newblock {\em arXiv preprint arXiv:2311.00571}, 2023.

\bibitem{dai2305instructBLIP}
W~Dai, J~Li, D~Li, AMH Tiong, J~Zhao, W~Wang, B~Li, P~Fung, and S~Hoi.
\newblock Instructblip: Towards general-purpose vision-language models with instruction tuning. arxiv 2023.
\newblock {\em arXiv preprint arXiv:2305.06500}, 2023.

\bibitem{ddupont808:2023:GPT-4V-Act}
ddupont808.
\newblock Gpt-4v-act.
\newblock \url{https://github.com/ddupont808/GPT-4V-Act}, 2023.
\newblock Accessed on \today.

\bibitem{fu2023drive}
Daocheng Fu, Xin Li, Licheng Wen, Min Dou, Pinlong Cai, Botian Shi, and Yu~Qiao.
\newblock Drive like a human: Rethinking autonomous driving with large language models.
\newblock {\em arXiv preprint arXiv:2307.07162}, 2023.

\bibitem{gao2023assistgui}
Difei Gao, Lei Ji, Zechen Bai, Mingyu Ouyang, Peiran Li, Dongxing Mao, Qinchen Wu, Weichen Zhang, Peiyi Wang, Xiangwu Guo, et~al.
\newblock Assistgui: Task-oriented desktop graphical user interface automation.
\newblock {\em arXiv preprint arXiv:2312.13108}, 2023.

\bibitem{gao2023assistgpt}
Difei Gao, Lei Ji, Luowei Zhou, Kevin~Qinghong Lin, Joya Chen, Zihan Fan, and Mike~Zheng Shou.
\newblock Assistgpt: A general multi-modal assistant that can plan, execute, inspect, and learn.
\newblock {\em arXiv preprint arXiv:2306.08640}, 2023.

\bibitem{gao2023clova}
Zhi Gao, Yuntao Du, Xintong Zhang, Xiaojian Ma, Wenjuan Han, Song-Chun Zhu, and Qing Li.
\newblock Clova: A closed-loop visual assistant with tool usage and update.
\newblock {\em arXiv preprint arXiv:2312.10908}, 2023.

\bibitem{gupta2023visual}
Tanmay Gupta and Aniruddha Kembhavi.
\newblock Visual programming: Compositional visual reasoning without training.
\newblock In {\em Proceedings of the IEEE/CVF Conference on Computer Vision and Pattern Recognition}, pages 14953--14962, 2023.

\bibitem{hu2023avis}
Ziniu Hu, Ahmet Iscen, Chen Sun, Kai-Wei Chang, Yizhou Sun, David~A Ross, Cordelia Schmid, and Alireza Fathi.
\newblock Avis: Autonomous visual information seeking with large language model agent.
\newblock In {\em Thirty-seventh Conference on Neural Information Processing Systems}, 2023.

\bibitem{huang2023audiogpt}
Rongjie Huang, Mingze Li, Dongchao Yang, Jiatong Shi, Xuankai Chang, Zhenhui Ye, Yuning Wu, Zhiqing Hong, Jiawei Huang, Jinglin Liu, et~al.
\newblock Audiogpt: Understanding and generating speech, music, sound, and talking head.
\newblock {\em arXiv preprint arXiv:2304.12995}, 2023.

\bibitem{karthik2023vision}
Shyamgopal Karthik, Karsten Roth, Massimiliano Mancini, and Zeynep Akata.
\newblock Vision-by-language for training-free compositional image retrieval.
\newblock {\em arXiv preprint arXiv:2310.09291}, 2023.

\bibitem{kirillov2023segment}
Alexander Kirillov, Eric Mintun, Nikhila Ravi, Hanzi Mao, Chloe Rolland, Laura Gustafson, Tete Xiao, Spencer Whitehead, Alexander~C Berg, Wan-Yen Lo, et~al.
\newblock Segment anything.
\newblock {\em arXiv preprint arXiv:2304.02643}, 2023.

\bibitem{Koh2024VisualWebArena}
Yujing Koh, Robert Lo, Lawrence Jang, Vikram Duvvur, Mingchong Lim, Po-yu Huang, Graham Neubig, Shuyan Zhou, Ruslan Salakhutdinov, and Daniel Fried.
\newblock Visualwebarena: Evaluating multimodal agents on realistic visual web tasks.
\newblock {\em arXiv preprint arXiv:2401.13649}, 2024.

\bibitem{lee2023explore}
Sunjae Lee, Junyoung Choi, Jungjae Lee, Hojun Choi, Steven~Y Ko, Sangeun Oh, and Insik Shin.
\newblock Explore, select, derive, and recall: Augmenting llm with human-like memory for mobile task automation.
\newblock {\em arXiv preprint arXiv:2312.03003}, 2023.

\bibitem{li2023BLIP}
Junnan Li, Dongxu Li, Silvio Savarese, and Steven Hoi.
\newblock Blip-2: Bootstrapping language-image pre-training with frozen image encoders and large language models.
\newblock {\em arXiv preprint arXiv:2301.12597}, 2023.

\bibitem{li2022BLIP}
Junnan Li, Dongxu Li, Caiming Xiong, and Steven Hoi.
\newblock Blip: Bootstrapping language-image pre-training for unified vision-language understanding and generation.
\newblock In {\em International Conference on Machine Learning}, pages 12888--12900. PMLR, 2022.

\bibitem{li2024mulan}
Sen Li, Ruochen Wang, Cho-Jui Hsieh, et~al.
\newblock Mulan: Multimodal-llm agent for progressive multi-object diffusion.
\newblock {\em arXiv preprint arXiv:2402.12741}, 2024.

\bibitem{li2022systematic}
Xiang~Lorraine Li, Adhiguna Kuncoro, Jordan Hoffmann, Cyprien de~Masson~d’Autume, Phil Blunsom, and Aida Nematzadeh.
\newblock A systematic investigation of commonsense knowledge in large language models.
\newblock In {\em Proceedings of the 2022 Conference on Empirical Methods in Natural Language Processing}, pages 11838--11855, 2022.

\bibitem{li2024styletts}
Yinghao~Aaron Li, Cong Han, Vinay Raghavan, Gavin Mischler, and Nima Mesgarani.
\newblock Styletts 2: Towards human-level text-to-speech through style diffusion and adversarial training with large speech language models.
\newblock {\em Advances in Neural Information Processing Systems}, 36, 2024.

\bibitem{liu2023llava}
Shilong Liu, Hao Cheng, Haotian Liu, Hao Zhang, Feng Li, Tianhe Ren, Xueyan Zou, Jianwei Yang, Hang Su, Jun Zhu, et~al.
\newblock Llava-plus: Learning to use tools for creating multimodal agents.
\newblock {\em arXiv preprint arXiv:2311.05437}, 2023.

\bibitem{liu2023grounding}
Shilong Liu, Zhaoyang Zeng, Tianhe Ren, Feng Li, Hao Zhang, Jie Yang, Chunyuan Li, Jianwei Yang, Hang Su, Jun Zhu, et~al.
\newblock Grounding dino: Marrying dino with grounded pre-training for open-set object detection.
\newblock {\em arXiv preprint arXiv:2303.05499}, 2023.

\bibitem{liu2023towards}
Xiangyan Liu, Rongxue Li, Wei Ji, and Tao Lin.
\newblock Towards robust multi-modal reasoning via model selection.
\newblock {\em arXiv preprint arXiv:2310.08446}, 2023.

\bibitem{liu2023wavjourney}
Xubo Liu, Zhongkai Zhu, Haohe Liu, Yi~Yuan, Meng Cui, Qiushi Huang, Jinhua Liang, Yin Cao, Qiuqiang Kong, Mark~D Plumbley, et~al.
\newblock Wavjourney: Compositional audio creation with large language models.
\newblock {\em arXiv preprint arXiv:2307.14335}, 2023.

\bibitem{liu2024multimodal}
Yang Liu, Xinshuai Song, Kaixuan Jiang, Weixing Chen, Jingzhou Luo, Guanbin Li, and Liang Lin.
\newblock Multimodal embodied interactive agent for cafe scene.
\newblock {\em arXiv preprint arXiv:2402.00290}, 2024.

\bibitem{liu2021paddleseg}
Yi~Liu, Lutao Chu, Guowei Chen, Zewu Wu, Zeyu Chen, Baohua Lai, and Yuying Hao.
\newblock Paddleseg: A high-efficient development toolkit for image segmentation.
\newblock {\em arXiv preprint arXiv:2101.06175}, 2021.

\bibitem{long2023discuss}
Yuxing Long, Xiaoqi Li, Wenzhe Cai, and Hao Dong.
\newblock Discuss before moving: Visual language navigation via multi-expert discussions.
\newblock {\em arXiv preprint arXiv:2309.11382}, 2023.

\bibitem{lu2023chameleon}
Pan Lu, Baolin Peng, Hao Cheng, Michel Galley, Kai-Wei Chang, Ying~Nian Wu, Song-Chun Zhu, and Jianfeng Gao.
\newblock Chameleon: Plug-and-play compositional reasoning with large language models.
\newblock {\em arXiv preprint arXiv:2304.09842}, 2023.

\bibitem{lu2024weblinx}
Xing~Han L{\`u}, Zden{\v{e}}k Kasner, and Siva Reddy.
\newblock Weblinx: Real-world website navigation with multi-turn dialogue.
\newblock {\em arXiv preprint arXiv:2402.05930}, 2024.

\bibitem{mao2023gpt}
Jiageng Mao, Yuxi Qian, Hang Zhao, and Yue Wang.
\newblock Gpt-driver: Learning to drive with gpt.
\newblock {\em arXiv preprint arXiv:2310.01415}, 2023.

\bibitem{maurer2016autonomous}
Markus Maurer, J~Christian Gerdes, Barbara Lenz, and Hermann Winner.
\newblock {\em Autonomous driving: technical, legal and social aspects}.
\newblock Springer Nature, 2016.

\bibitem{mialon2023gaia}
Gr{\'e}goire Mialon, Cl{\'e}mentine Fourrier, Craig Swift, Thomas Wolf, Yann LeCun, and Thomas Scialom.
\newblock Gaia: a benchmark for general ai assistants.
\newblock {\em arXiv preprint arXiv:2311.12983}, 2023.

\bibitem{osoba2020policy}
Osonde~A Osoba, Raffaele Vardavas, Justin Grana, Rushil Zutshi, and Amber Jaycocks.
\newblock Policy-focused agent-based modeling using rl behavioral models.
\newblock {\em arXiv preprint arXiv:2006.05048}, 2020.

\bibitem{pan2023large}
Jeff~Z Pan, Simon Razniewski, Jan-Christoph Kalo, Sneha Singhania, Jiaoyan Chen, Stefan Dietze, Hajira Jabeen, Janna Omeliyanenko, Wen Zhang, Matteo Lissandrini, et~al.
\newblock Large language models and knowledge graphs: Opportunities and challenges.
\newblock {\em arXiv preprint arXiv:2308.06374}, 2023.

\bibitem{qin2023mp5}
Yiran Qin, Enshen Zhou, Qichang Liu, Zhenfei Yin, Lu~Sheng, Ruimao Zhang, Yu~Qiao, and Jing Shao.
\newblock Mp5: A multi-modal open-ended embodied system in minecraft via active perception.
\newblock {\em arXiv preprint arXiv:2312.07472}, 2023.

\bibitem{qin2023toolllm}
Yujia Qin, Shihao Liang, Yining Ye, Kunlun Zhu, Lan Yan, Yaxi Lu, Yankai Lin, Xin Cong, Xiangru Tang, Bill Qian, et~al.
\newblock Toolllm: Facilitating large language models to master 16000+ real-world apis.
\newblock {\em arXiv preprint arXiv:2307.16789}, 2023.

\bibitem{rombach2022high}
Robin Rombach, Andreas Blattmann, Dominik Lorenz, Patrick Esser, and Bjorn Ommer.
\newblock High-resolution image synthesis with latent diffusion models.
\newblock In {\em Proceedings of the IEEE/CVF conference on computer vision and pattern recognition}, pages 10684--10695, 2022.

\bibitem{schick2023toolformer}
Timo Schick, Jane Dwivedi-Yu, Roberto Dess{\`\i}, Roberta Raileanu, Maria Lomeli, Luke Zettlemoyer, Nicola Cancedda, and Thomas Scialom.
\newblock Toolformer: Language models can teach themselves to use tools.
\newblock {\em arXiv preprint arXiv:2302.04761}, 2023.

\bibitem{shen2023hugginggpt}
Yongliang Shen, Kaitao Song, Xu~Tan, Dongsheng Li, Weiming Lu, and Yueting Zhuang.
\newblock Hugginggpt: Solving ai tasks with chatgpt and its friends in huggingface.
\newblock {\em arXiv preprint arXiv:2303.17580}, 2023.

\bibitem{sumers2023cognitive}
Theodore Sumers, Shunyu Yao, Karthik Narasimhan, and Thomas~L Griffiths.
\newblock Cognitive architectures for language agents.
\newblock {\em arXiv preprint arXiv:2309.02427}, 2023.

\bibitem{suris2023vipergpt}
D{\'\i}dac Sur{\'\i}s, Sachit Menon, and Carl Vondrick.
\newblock Vipergpt: Visual inference via python execution for reasoning.
\newblock {\em arXiv preprint arXiv:2303.08128}, 2023.

\bibitem{tao2023webwise}
Heyi Tao, Sethuraman TV, Michal Shlapentokh-Rothman, Derek Hoiem, and Heng Ji.
\newblock Webwise: Web interface control and sequential exploration with large language models.
\newblock {\em arXiv preprint arXiv:2310.16042}, 2023.

\bibitem{vemprala2023grid}
Sai Vemprala, Shuhang Chen, Abhinav Shukla, Dinesh Narayanan, and Ashish Kapoor.
\newblock Grid: A platform for general robot intelligence development.
\newblock {\em arXiv preprint arXiv:2310.00887}, 2023.

\bibitem{Wang2024MLLM-Tool}
Chenyu Wang, Weixin Luo, Qianyu Chen, Haonan Mai, jindi Guo, Sixun Dong, Xiaohua Xuan, Zhengxin Li, Lin Ma, and Shenghua Gao.
\newblock Mllm-tool: A multimodal large language model for tool agent learning.
\newblock {\em arXiv preprint arXiv:2401.10727}, 2024.

\bibitem{wang2023chatvideo}
Junke Wang, Dongdong Chen, Chong Luo, Xiyang Dai, Lu~Yuan, Zuxuan Wu, and Yu-Gang Jiang.
\newblock Chatvideo: A tracklet-centric multimodal and versatile video understanding system.
\newblock {\em arXiv preprint arXiv:2304.14407}, 2023.

\bibitem{wang2024mobile}
Junyang Wang, Haiyang Xu, Jiabo Ye, Ming Yan, Weizhou Shen, Ji~Zhang, Fei Huang, and Jitao Sang.
\newblock Mobile-agent: Autonomous multi-modal mobile device agent with visual perception.
\newblock {\em arXiv preprint arXiv:2401.16158}, 2024.

\bibitem{wang2023survey}
Lei Wang, Chen Ma, Xueyang Feng, Zeyu Zhang, Hao Yang, Jingsen Zhang, Zhiyuan Chen, Jiakai Tang, Xu~Chen, Yankai Lin, et~al.
\newblock A survey on large language model based autonomous agents.
\newblock {\em arXiv preprint arXiv:2308.11432}, 2023.

\bibitem{wang2020completely}
Xiaoling Wang and Housheng Su.
\newblock Completely model-free rl-based consensus of continuous-time multi-agent systems.
\newblock {\em Applied Mathematics and Computation}, 382:125312, 2020.

\bibitem{wang2023jarvis}
Zihao Wang, Shaofei Cai, Anji Liu, Yonggang Jin, Jinbing Hou, Bowei Zhang, Haowei Lin, Zhaofeng He, Zilong Zheng, Yaodong Yang, et~al.
\newblock Jarvis-1: Open-world multi-task agents with memory-augmented multimodal language models.
\newblock {\em arXiv preprint arXiv:2311.05997}, 2023.

\bibitem{wang2023describe}
Zihao Wang, Shaofei Cai, Anji Liu, Xiaojian Ma, and Yitao Liang.
\newblock Describe, explain, plan and select: Interactive planning with large language models enables open-world multi-task agents.
\newblock {\em arXiv preprint arXiv:2302.01560}, 2023.

\bibitem{wen2023empowering}
Hao Wen, Yuanchun Li, Guohong Liu, Shanhui Zhao, Tao Yu, Toby Jia-Jun Li, Shiqi Jiang, Yunhao Liu, Yaqin Zhang, and Yunxin Liu.
\newblock Empowering llm to use smartphone for intelligent task automation.
\newblock {\em arXiv preprint arXiv:2308.15272}, 2023.

\bibitem{wen2023droidbot}
Hao Wen, Hongming Wang, Jiaxuan Liu, and Yuanchun Li.
\newblock Droidbot-gpt: Gpt-powered ui automation for android.
\newblock {\em arXiv preprint arXiv:2304.07061}, 2023.

\bibitem{wen2023road}
Licheng Wen, Xuemeng Yang, Daocheng Fu, Xiaofeng Wang, Pinlong Cai, Xin Li, Tao Ma, Yingxuan Li, Linran Xu, Dengke Shang, et~al.
\newblock On the road with gpt-4v (ision): Early explorations of visual-language model on autonomous driving.
\newblock {\em arXiv preprint arXiv:2311.05332}, 2023.

\bibitem{wooldridge1995intelligent}
Michael Wooldridge and Nicholas~R Jennings.
\newblock Intelligent agents: Theory and practice.
\newblock {\em The knowledge engineering review}, 10(2):115--152, 1995.

\bibitem{wu2023visual}
Chenfei Wu, Shengming Yin, Weizhen Qi, Xiaodong Wang, Zecheng Tang, and Nan Duan.
\newblock Visual chatgpt: Talking, drawing and editing with visual foundation models.
\newblock {\em arXiv preprint arXiv:2303.04671}, 2023.

\bibitem{wu2023smartplay}
Yue Wu, Xuan Tang, Tom~M Mitchell, and Yuanzhi Li.
\newblock Smartplay: A benchmark for llms as intelligent agents.
\newblock {\em arXiv preprint arXiv:2310.01557}, 2023.

\bibitem{wu2024copilot}
Zhiyong Wu, Chengcheng Han, Zichen Ding, Zhenmin Weng, Zhoumianze Liu, Shunyu Yao, Tao Yu, and Lingpeng Kong.
\newblock Os-copilot: Towards generalist computer agents with self-improvement.
\newblock {\em arXiv preprint arXiv:2402.07456}, 2024.

\bibitem{xi2023rise}
Zhiheng Xi, Wenxiang Chen, Xin Guo, Wei He, Yiwen Ding, Boyang Hong, Ming Zhang, Junzhe Wang, Senjie Jin, Enyu Zhou, et~al.
\newblock The rise and potential of large language model based agents: A survey.
\newblock {\em arXiv preprint arXiv:2309.07864}, 2023.

\bibitem{xie2024travelplanner}
Jian Xie, Kai Zhang, Jiangjie Chen, Tinghui Zhu, Renze Lou, Yuandong Tian, Yanghua Xiao, and Yu~Su.
\newblock Travelplanner: A benchmark for real-world planning with language agents.
\newblock {\em arXiv preprint arXiv:2402.01622}, 2024.

\bibitem{xie2023openagents}
Tianbao Xie, Fan Zhou, Zhoujun Cheng, Peng Shi, Luoxuan Weng, Yitao Liu, Toh~Jing Hua, Junning Zhao, Qian Liu, Che Liu, et~al.
\newblock Openagents: An open platform for language agents in the wild.
\newblock {\em arXiv preprint arXiv:2310.10634}, 2023.

\bibitem{xu2023instructp2p}
Jiale Xu, Xintao Wang, Yan-Pei Cao, Weihao Cheng, Ying Shan, and Shenghua Gao.
\newblock Instructp2p: Learning to edit 3d point clouds with text instructions.
\newblock {\em arXiv preprint arXiv:2306.07154}, 2023.

\bibitem{yan2023gpt}
An~Yan, Zhengyuan Yang, Wanrong Zhu, Kevin Lin, Linjie Li, Jianfeng Wang, Jianwei Yang, Yiwu Zhong, Julian McAuley, Jianfeng Gao, et~al.
\newblock Gpt-4v in wonderland: Large multimodal models for zero-shot smartphone gui navigation.
\newblock {\em arXiv preprint arXiv:2311.07562}, 2023.

\bibitem{yang2023octopus}
Jingkang Yang, Yuhao Dong, Shuai Liu, Bo~Li, Ziyue Wang, Chencheng Jiang, Haoran Tan, Jiamu Kang, Yuanhan Zhang, Kaiyang Zhou, et~al.
\newblock Octopus: Embodied vision-language programmer from environmental feedback.
\newblock {\em arXiv preprint arXiv:2310.08588}, 2023.

\bibitem{yang2023supervised}
Linyi Yang, Shuibai Zhang, Zhuohao Yu, Guangsheng Bao, Yidong Wang, Jindong Wang, Ruochen Xu, Wei Ye, Xing Xie, Weizhu Chen, et~al.
\newblock Supervised knowledge makes large language models better in-context learners.
\newblock {\em arXiv preprint arXiv:2312.15918}, 2023.

\bibitem{yang2023gpt4tools}
Rui Yang, Lin Song, Yanwei Li, Sijie Zhao, Yixiao Ge, Xiu Li, and Ying Shan.
\newblock Gpt4tools: Teaching large language model to use tools via self-instruction.
\newblock {\em arXiv preprint arXiv:2305.18752}, 2023.

\bibitem{yang2023embodied}
Yijun Yang, Tianyi Zhou, Kanxue Li, Dapeng Tao, Lusong Li, Li~Shen, Xiaodong He, Jing Jiang, and Yuhui Shi.
\newblock Embodied multi-modal agent trained by an llm from a parallel textworld.
\newblock {\em arXiv preprint arXiv:2311.16714}, 2023.

\bibitem{yang2023appagent}
Zhao Yang, Jiaxuan Liu, Yucheng Han, Xin Chen, Zebiao Huang, Bin Fu, and Gang Yu.
\newblock Appagent: Multimodal agents as smartphone users.
\newblock {\em arXiv preprint arXiv:2312.13771}, 2023.

\bibitem{yang2023mm}
Zhengyuan Yang, Linjie Li, Jianfeng Wang, Kevin Lin, Ehsan Azarnasab, Faisal Ahmed, Zicheng Liu, Ce~Liu, Michael Zeng, and Lijuan Wang.
\newblock Mm-react: Prompting chatgpt for multimodal reasoning and action.
\newblock {\em arXiv preprint arXiv:2303.11381}, 2023.

\bibitem{Yang2024DoraemonGPT}
Zongxin Yang, Guikun Chen, Xiaodi Li, Wenguan Wang, and Yi~Yang.
\newblock Doraemongpt: Toward understanding dynamic scenes with large language models.
\newblock {\em arXiv preprint arXiv:2401.08392}, 2024.

\bibitem{ye2022joint}
Botao Ye, Hong Chang, Bingpeng Ma, Shiguang Shan, and Xilin Chen.
\newblock Joint feature learning and relation modeling for tracking: A one-stream framework.
\newblock In {\em European Conference on Computer Vision}, pages 341--357. Springer, 2022.

\bibitem{yu2023musicagent}
Dingyao Yu, Kaitao Song, Peiling Lu, Tianyu He, Xu~Tan, Wei Ye, Shikun Zhang, and Jiang Bian.
\newblock Musicagent: An ai agent for music understanding and generation with large language models.
\newblock {\em arXiv preprint arXiv:2310.11954}, 2023.

\bibitem{yuan2023CRAFT}
Lifan Yuan, Yangyi Chen, Xingyao Wang, Yi~R. Fung, Hao Peng, and Heng Ji.
\newblock Craft: Customizing llms by creating and retrieving from specialized toolsets.
\newblock {\em arXiv preprint arXiv:2309.17428}, 2023.

\bibitem{zhan2023you}
Zhuosheng Zhan and Aston Zhang.
\newblock You only look at screens: Multimodal chain-of-action agents.
\newblock {\em arXiv preprint arXiv:2309.11436}, 2023.

\bibitem{zhang2023bootstrap}
Jesse Zhang, Jiahui Zhang, Karl Pertsch, Ziyi Liu, Xiang Ren, Minsuk Chang, Shao-Hua Sun, and Joseph~J Lim.
\newblock Bootstrap your own skills: Learning to solve new tasks with large language model guidance.
\newblock {\em arXiv preprint arXiv:2310.10021}, 2023.

\bibitem{zhang2023loop}
Yixiao Zhang, Akira Maezawa, Gus Xia, Kazuhiko Yamamoto, and Simon Dixon.
\newblock Loop copilot: Conducting ai ensembles for music generation and iterative editing.
\newblock {\em arXiv preprint arXiv:2310.12404}, 2023.

\bibitem{zhang2023large}
Zihan Zhang, Meng Fang, Ling Chen, Mohammad-Reza Namazi-Rad, and Jun Wang.
\newblock How do large language models capture the ever-changing world knowledge? a review of recent advances.
\newblock {\em arXiv preprint arXiv:2310.07343}, 2023.

\bibitem{zhao2023see}
Zhonghan Zhao, Wenhao Chai, Xuan Wang, Li~Boyi, Shengyu Hao, Shidong Cao, Tian Ye, Jenq-Neng Hwang, and Gaoang Wang.
\newblock See and think: Embodied agent in virtual environment.
\newblock {\em arXiv preprint arXiv:2311.15209}, 2023.

\bibitem{zheng2023ddcot}
Ge~Zheng, Bin Yang, Jiajin Tang, Hong-Yu Zhou, and Sibei Yang.
\newblock Ddcot: Duty-distinct chain-of-thought prompting for multimodal reasoning in language models.
\newblock {\em arXiv preprint arXiv:2310.16436}, 2023.

\bibitem{zhou2023vision}
Xingcheng Zhou, Mingyu Liu, Bare~Luka Zagar, Ekim Yurtsever, and Alois~C Knoll.
\newblock Vision language models in autonomous driving and intelligent transportation systems.
\newblock {\em arXiv preprint arXiv:2310.14414}, 2023.

\end{thebibliography}

\end{document}